# Prosper: image and robot-guided prostate brachytherapy

Michael Baumann[a], Michel Bolla[b], Vincent Daanen[c], Jean-Luc Descotes[d], Jean-Yves Giraud[b], Nikolai Hungr[a], Antoine Leroy[c], Jean-Alexandre Long[a,d], Sébastien Martin[a], Jocelyne Troccaz[1,a,d]

[a]UJF-Grenoble I/CNRS/TIMC-IMAG laboratory (UMR5525), Faculté de Médecine, Domaine de la Merci, 38706 La Tronche, France
[b]Radiation oncology department, Grenoble University Hospital, BP 217, Grenoble cedex 09, France
[c]Koelis, 5, Avenue du Grand Sablon, 38700 La Tronche, France
[d]Urology department, Grenoble University Hospital, BP 217, Grenoble cedex 09, France

Résumé : La curiethérapie pour cancer localisé de la prostate consiste à insérer dans la glande, au moyen d'aiguilles implantées de façon transpérinéale, des grains d'iode radioactifs (I125) pour le détruire. L'insertion des grains après planification préalable de leur position se fait sous contrôle échographique per-opératoire. Nous proposons d'optimiser la procédure par : i) le recours à une imagerie ultrasonore 3D, ii) l'introduction de données IRM par fusion avec l'imagerie ultrasonore. iii) l'utilisation d'un robot spécifiquement conçu pour l'insertion des aiguilles, connecté à l'imagerie. Les procédures d'imagerie implémentées ont été testées avec succès sur des données de patients et la précision du robot a été évaluée sur fantôme déformable réaliste.

Abstract: Brachytherapy for localized prostate cancer consists in destroying cancer by introducing iodine radioactive seeds into the gland through hollow needles. The planning of the position of the seeds and their introduction into the prostate is based on intra-operative ultrasound (US) imaging. We propose to optimize the global quality of the procedure by: i) using 3D US; ii) enhancing US data with MRI registration; iii) using a specially designed needle-insertion robot, connected to the imaging data. The imaging methods have been successfully tested on patient data while the robot accuracy has been evaluated on a realistic deformable phantom.

Introduction:
Prostate cancer is the second most frequent cancer for men in many developed countries. For localized prostate cancer, different treatments are proposed depending on clinical stage, Gleason score and baseline PSA. Together with radical prostatectomy and intensity modulated radiotherapy (IMRT), brachytherapy is one of the proposed treatments. It consists in placing radioactive seeds in the prostate for cancer destruction. These seeds are inserted through hollow needles. Planning and seed insertion are based on US images most often acquired by a conventional 2D B-mode endorectal probe mounted on a stepper for horizontal sweeping of the prostate volume. Needles are inserted through a template that offers horizontal, parallel trajectories in a five millimeter vertical and horizontal grid (see figure 1). This procedure may face several difficulties: one comes from a possible conflict between the horizontal needle trajectory and the pubic arch (for large prostates in particular); another is due to the fact that the prostate moves and can be deformed during needle insertion or from US probe motion; prostate edema can also significantly modify the prostate volume (up to 20%) during the procedure. These potential difficulties may result in inaccurate placement of the seeds.

The Prosper project aims at reducing the inaccuracies of seed placement. To make planning easier, we propose [1] to enhance US data by the addition of pre-operative MRI information registered to the US images. To limit the deformation of the prostate due to US probe motion during volume collection, we replaced the 2D probe and stepper by a stationary 3D US probe. Finally, to allow for a larger variety of needle trajectories and to limit the motion and deformation of the prostate during needle insertion, we developed a special needle-insertion robot. Seed insertion through the needles is still manual. In the following sections, we describe the imaging and robotic components of Prosper, along with the experiments conducted, their results and future work.

---

[1]Author for correspondence: Jocelyne.Troccaz@imag.fr

Materials and methods:
As mentioned above, two imaging modalities are used in Prosper: US and MRI T2. Special efforts have been made to limit the user involvement in image processing tasks, especially prostate segmentation in the images. To automate MRI segmentation of data, an atlas combining statistics of the shape of the prostate and statistics of its appearance in MRI data was computed from a set of MRI exams manually segmented by an expert. This allows for the segmentation of new MRI volumes with user interaction limited to localized corrections. This work is described in [2]. Regarding US segmentation, we developed both a semi-interactive approach based on the fitting of the average prostate of the statistical model to a few points given by the user, as well as a fully automatic approach combining an appearance model of the prostate in US data learnt from examples, to the prostate segmented in the MRI data. MRI and US contours are finally registered using volume constraints and an elastic deformation model. This is described in [3].

The robot consists of two main components; one module allows placing and orienting the needle at the entry point close to the perineum. It consists of 5 degrees of freedom. The second module is for needle insertion: it combines a translation degree of freedom with an axial rotation of the needle during insertion. The latter helps to limit tissue deformation during needle insertion. A passive mechanism has been added to stop the motion in case of conflict with the pubic arch. The robot architecture is described in [4].

These components are illustrated in figure 2.

Results:
Segmentation methods have been tested on data coming from real patients (see [2,3]). The accuracy, as compared to manual segmentation, is most often very good and the user always has the ability to make local corrections when necessary. The main advantage comes from the limited time spent by the user during the segmentation process compared to the labor-intensive fully manual process.

The robot has been calibrated with respect to the 3D US probe using a set of needle insertions detected in a water phantom. The robot is able to reach a target in the water with a precision of less than 1 mm throughout the entire workspace. A specific phantom has also been designed for robot performance evaluation. This phantom includes a rectum, a prostate with different layers and a perineum. The prostate inside the phantom is realistically mobile and deformable. The phantom can be imaged by US, CT and MRI; fiducials for accuracy evaluation can also be included. Experiments with this phantom are in progress. We can reasonably expect an accuracy of less than 2 mm.

Discussion:
A prototype of an image and robot-guided brachytherapy system has been designed and successfully evaluated. Industrial versions of the imaging modules have now been developed and are being integrated into products. From the clinical point of view, a slightly modified version of the robot must be designed before testing on patients. The current version, a laboratory prototype, does not fulfill regulation requirements yet. From an academic point of view, future work will deal with the automatic tracking of the prostate and seeds from 3D intra-operative US volumes. Finally, the system could be adapted for other applications requiring perineal access to the prostate: for instance biopsies, placement of probes for HIFU or interstitial laser treatment.

Acknowledgments: This research project was financially supported by the French National Research Agency (ANR) through the call for proposals TecSan (Prosper project) and by Université Joseph Fourier (UJF-Grenoble I).

References:
[1] Daanen V, Gastaldo J, Giraud J-Y, Fourneret P, Descotes J-L, Bolla M, Collomb D, Troccaz J. MRI/TRUS data fusion for brachytherapy. The International Journal of Medical Robotics and Computer-Assisted Surgery, Vol2, No.3, pp256-261, September 2006
[2] Martin S, Troccaz J, Daanen V. Automated Segmentation of the Prostate in 3D MR Images Using a Probabilistic Atlas and a Spatially Constrained Deformable Model. Medical Physics, 2010, 37(4):1579-1590
[3] Martin S, Baumann M, Daanen V, Troccaz J. MR prior based automatic segmentation of the prostate in TRUS images for MR/TRUS data fusion. IEEE International Symposium on Biomedical Imaging, ISBI'2010, Rotterdam, 14-17 Avril 2010, pp640-643
[4] Hungr N, Troccaz J, Zemiti N, Tripodi N. Design of an Ultrasound-Guided Robotic Brachytherapy Needle-Insertion System. Proceedings of IEEE EMBC'2009, pp250-253, Minneapolis, 2-6 septembre 2009

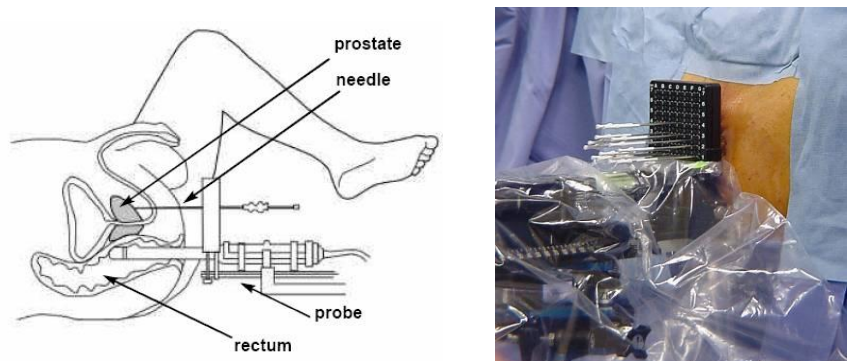

Figure 1: Prostate brachytherapy set-up
(left: from http://www.prostatebrachytherapyinfo.net)

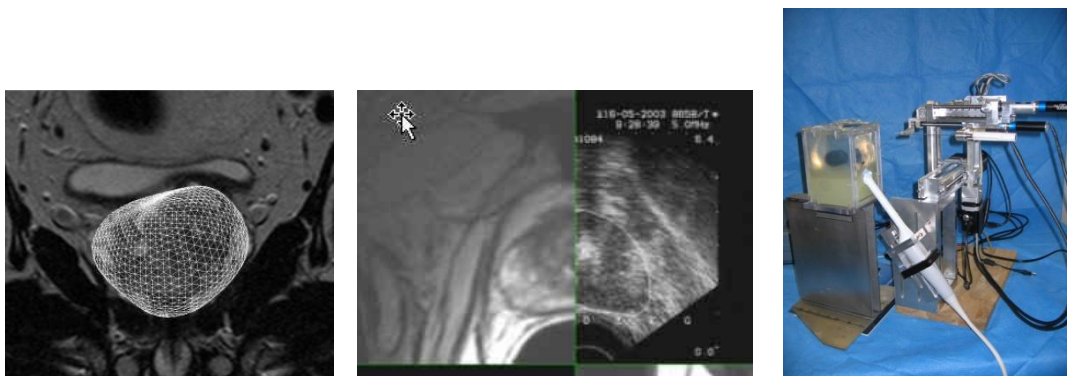

Figure 2: Prosper components
From left to right: atlas-based segmentation of MRI data, US-to-MRI registration, experimental set-up (robot, 3D US probe and deformable phantom)